# Lightweight Spatial-Channel Adaptive Coordination of Multilevel Refinement Enhancement Network for Image Reconstruction

Yuxi Cai, Huicheng Lai*, Zhenhong Jia

College of Information Science and Engineering, Xinjiang University, Urumqi 830046, China

Benefiting from the vigorous development of deep learning, many CNN-based image super-resolution methods have emerged and achieved better results than traditional algorithms. However, it is difficult for most algorithms to adaptively adjust the spatial region and channel features at the same time, let alone the information exchange between them. In addition, the exchange of information between attention modules is even less visible to researchers. To solve these problems, we put forward a lightweight spatial-channel adaptive coordination of multilevel refinement enhancement networks(MREN). Specifically, we construct a space-channel adaptive coordination block, which enables the network to learn the spatial region and channel feature information of interest under different receptive fields. In addition, the information of the corresponding feature processing level between the spatial part and the channel part is exchanged with the help of jump connection to achieve the coordination between the two. We establish a communication bridge between attention modules through a simple linear combination operation, so as to more accurately and continuously guide the network to pay attention to the information of interest. Extensive experiments on several standard test sets have shown that our MREN achieves superior performance over other advanced algorithms with a very small number of parameters and very low computational complexity.

## 1. Introduction

Image super-resolution reconstruction is one of the classic underlying computer vision tasks that have been widely concerned. It aims to use certain technical ways to reconstruct visually pleasant high-resolution images(HR) from degraded low-resolution images(LR). It has been widely used in remote sensing[SwinSUNet: Pure Transformer Network for Remote Sensing Image Change Detection], communication[Offset Learning Based Channel Estimation for Intelligent Reflecting Surface-Assisted Indoor Communication], medicine [A new sparse representation framework for compressed sensing MRI]、[Joint low-rank prior and difference of Gaussian filter for magnetic resonance image denoising]、[A Novel Iterative Shrinkage Algorithm for CS-MRI via Adaptive Regularization]and other fields. In recent years, with the rapid development of deep learning, the reconstruction method based on deep learning has been widely concerned by researchers. With the powerful nonlinear characterization ability of convolution neural networks, researchers have proposed many reconstruction algorithms with excellent performance, such as DNCL[1], FilterNet[2], ESRGCNN[3], etc. In 2016, Dong[4] constructed a shallow neural network containing three convolution layers to learn the mapping relationship between LR and HR, and obtained more excellent reconstruction results than traditional algorithms. Kim[5] constructed VDSR containing 20 layers with the power of residual learning, which solved the problem of difficult training of the network due to the deepening of the network, and the network performance was further improved. Tai[6], Liu[7] use dense connections at the end of the basic block to aggregate the hierarchical features of different stages, and the network can make use of the hierarchical features of each stage. Subsequently, the EDSR proposed by Lim[8] pushes the depth of the network to more than 160 layers. Tian[9] mitigated the network performance degradation problem due to up-sampling by a simple but effective cascaded network structure in CFSRCNN. Luo[10] designed a lattice filter based on butterfly structure and integrated context information through reverse feature fusion strategy. Under the condition of sacrificing certain computational complexity and memory storage, the network achieved good reconstruction results.

Although good reconstruction performance can be obtained by building a very deep networks, it also exposes the networks to problems such as large storage overhead and computational consumption, which makes them difficult to apply to the small removable devices with limited storage and computational power. From this perspective, it is particularly important to design lightweight and efficient reconstruction algorithms. Kim[11] constructed DRCN to reduce network' parameters by the recurrent learning shared parameters, and similarly DRRN[12]. However, these recursive methods only bring limited performance improvement to the network. SMSR[13] generates spatial and channel masks by sparse mask module to locate redundant computations, and then dynamically skips redundant computations with the help of sparse mask convolution to achieve efficient image SR. Ahn[14] uses cascading mechanisms at the local and global levels to integrate features from multiple phases while keeping the network lightweight. Lan[15], He[16] obtained the channel information under different receptive fields by several parallel branches separately, which allows the network to adaptively adjust the learning of important channel features. However, the lack of effective information exchange between branches makes it difficult for the network to work more coordinated and efficient. Cai[17] deals with different numbers of features by means of multiple branches with attention modules, and adaptively selects important features of different stages through splicing and convolution operations, but the network ignores the adaptation in the spatial area. Liu[18] also uses multiple branches in the basic block to obtain hierarchical features, but there is a lack of information exchange between branches and the network can not adaptively learn the spatial areas of interest. Although these networks achieve good reconstructions while remaining lightweight, they only achieve adaptive selection on the channels but do not allow the network to learn the local spatial information of the features more freely. Not to mention, let the network take into account both spatial adaptability and channel adaptability. In addition, these networks use attention module such as channel attention module or space attention module independently in the base block, and the lack of effective information exchange between neighboring attention modules creates a discontinuous flow of attention information and fails to continuously direct the network's attention to the information of interest. Finally, it is possible to further reduce the number of network parameters while maintaining the performance of the network.

To solve these problems, we propose a lightweight spatial-channel adaptive coordination of multilevel refinement enhancement networks for image reconstruction, which mainly consists of Multilevel Refinement Enhancement Block(MREB), Space-Channel Adaptive Coordination Block(SCACB), Double Refinement Attention Communication Block(DRACB), Reconstruction Block With Attention(RBWA) and so on. In detail, we divide SCACB into a spatial adaptive part and a channel adaptive part to learn spatial area information and channel feature information under different receptive fields, respectively. In order to coordinate and efficiently guide the network to learn more powerful feature representation, we exchange information in the corresponding feature processing stage of space part and channel part with the help of simple jump connection, and realize the interactive learning of spatial region and channel features. Then the spatial and channel information is aggregated step by step. MREB uses SCACB as the backbone for feature processing and double refinement of features through DRACB to achieve the preservation of important features in each stage and reduce the continuous processing of redundant information in the network. Most importantly, there will be attention information communication between adjacent DRACB, which can not only maintain the continuous transmission of attention information, but also adaptively adjust the network's attention to key information, so as to better guide the network to represent the important features. The first and second parts of this paper are the introduction and related work, and the third part will introduce the overall flow of the algorithm and the important components of the network in detail. The fourth part describes in detail the experimental settings and the ablation research of the algorithm, as well as the comparison with the results of other advanced algorithms. The fifth part is the conclusion. Our contribution mainly includes the following points:

1. The space-channel adaptive coordination block(SCACB) is proposed, which cleverly combines the learning of spatial information and channel information of features by the network. The SCACB can adaptively learn spatial region information and channel feature information under different receptive fields. Moreover, with the help of jump connections, SCACB achieves the information fusion of the features of the corresponding phases at the spatial and channel levels, which promotes its more coordinated and efficient work.

2. The double refinement attention communication block (DRACB) is constructed, which achieves both feature compression and attention feature map generation through 1×1 convolution layers. The most important and special thing is that we realize the communication of attention information between adjacent attention module through a simple linear combination operation, and maintain the continuous transmission of attention information, so as to better continuously guide the network to pay attention to the information of interest.

3. Combined with multilevel refinement enhancement block (MREB), we propose MREN, which achieves very competitive reconstruction results while keeping the network extremely lightweight and low computational complexity.

## 2. Related work

Group convolution [SRGC-Nets: Sparse Repeated Group Convolutional Neural Networks], depth-wise separable convolutions [A deep translation (GAN) based change detection network for optical and SAR remote sensing images], and self-calibrated convolution have been proposed to accelerate the deep models. Although these convolution types are more lightweight compared to ordinary convolution, it is difficult to fully utilize the representational power of convolution neural networks. In addition, it is important to think about how to effectively aggregate the hierarchical features at each stage to form a rich variety of feature libraries from which the network can draw its power. WMRN [19], MADNet [20] use convolution layers with different void rates in multiple branches to obtain channel information under different perceptual fields, and then use jump joints or splicing to aggregate features at each level. Similarly, LMAN[21]. Hui[22], Hui[23] used channel splitting operation in the base block to distill important features at different stages and then aggregated these features by various means. Based on this, Liu[24] changed the distillation method and used more flexible 1 × 1 convolution layers to distill out the key features at different stages, and the reconstruction performance was further improved. Wang[25] used a similar distillation mechanism in the base block and also obtained good improvement. However, these algorithms perform feature compression brutally, which can result in the loss of some key information. To this end, Cai[26] improved the distillation method, gradually distilling the key features of each stage from coarse to fine, and then using convolution layer to polymerize these features. Wang[13] proposed a scheme for adaptive learning of airspace and channel masks, which greatly reduces the computational complexity of the network while maintaining the performance of the network. Gao[27] proposed wide residual distillation connection, which connects the features of different convolution stages in the module while maintaining the light weight of the network, so as to realize the information exchange of different scale features. Tian[3] alternately uses ordinary convolution and lightweight grouped convolution to achieve feature extraction and multi-stage feature aggregation. In addition, [28], [29], [30] also achieve excellent reconstruction performance with a small number of parameters and computational complexity.

Many studies have shown that attention mechanisms play a crucial role in improving the performance of networks. SENet[31] can adaptively correct the interdependence of features. Hui[23] proposes contrast-aware channel attention module by using the sum of the global average and standard deviation of features to characterize channel characteristics, which makes the network pay more attention to high-frequency information in low-frequency space. Although channel attention improves the sensitivity of the network to key features, it ignores the importance of location information in generating the attention selection graph. Hou[32] proposed coordinate attention, which embeds the location information of features into channel attention through pooling operations, allowing the network to aggregate features along horizontal and vertical directions. Niu[33] designed a layer attention block and a channel space attention block to more comprehensively and selectively exploit information-rich features by modeling the inter-dependencies between different layers, channels, and locations. Cai[26] proposed a weight-sharing information lossless attention block, which enhances the recovery of high-frequency information such as edge textures through a weight-sharing auxiliary branching module. Wang[34] constructed a lightweight recurrent residual channel attention block to further improve the network performance by introducing recurrent connections in the attention module with a reduced number of module parameters. Zhao[35] proposed a pixel attention that can generate 3D attention maps to raise the attention level of the network to the pixel level.

## 3. Proposed Method

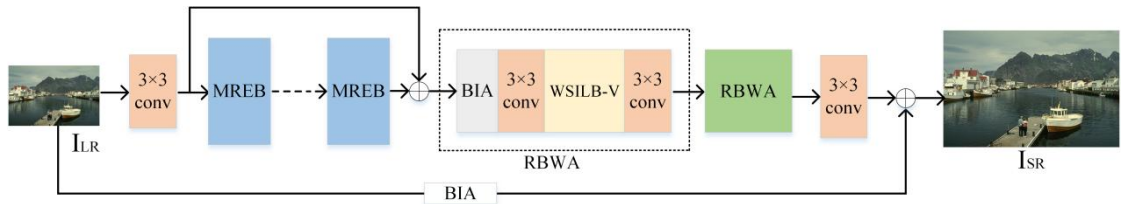

Figure 1 Overall block diagram of the network (MREB represents Multilevel Refinement Enhancement Block, BIA represents Bilinear Interpolation Algorithm, RBWA represents Reconstruction Block With Attention)

In this section, we will first introduce the overall flow of the algorithm in this paper, and then detail each important component of the network separately.

### 3.1 Network Architecture

The overall network block diagram of the MREN we designed is shown in figure 1, which is mainly composed of three parts: shallow feature extraction block, multilevel refinement enhancement block(MREB) and reconstruction block with attention(RBWA). In this paper, $I_{LR}$ and $I_{SR}$ are represented as the input and output of the network respectively. According to the research of [36] and [37], we only use a 3 × 3 convolution layer as the shallow feature extraction block to extract the shallow features:

$$F_0 = Conv_{3\times3}(I_{LR}) \tag{1}$$

Where $F_0$ represents the shallow features acquired by the network, and $Conv_{3\times3}(\cdot)$ denotes the convolution operation with a convolution kernel of 3 × 3. Then, the $F_0$ is transported to multiple cascaded MREB for deep processing of the features. Suppose there are n MREB, and the output features of the nth MREB is expressed as follows:

$$F_n = H_{MREB,n}(F_{n-1}) = H_{MREB,n}(H_{MREB,n-1}(\ldots(H_{MREB,1}(F_0))\ldots)) \tag{2}$$

Here, $H_{MREB,n}(\cdot)$ represents the nth MREB, $F_n$ represents the deep features acquired after the nth MREB processing, and more details about MREB will be described in 3.4.

After several MREB processing, we have learned the distinguishing deep features that contain rich high-frequency information, but in the continuous deepening processing of the features, the network has lost some of the underlying information in the low-resolution space, and this information plays an important role in the reconstruction performance of the network. To this end, we have carried out a jump spread of $F_0$. Then, $F_n$ is restored to the corresponding target size by the reconstruction block with attention(RBWA):

$$F_m = H_{RBWA}^m(H_{RBWA}^{m-1}(\cdots H_{RBWA}^0(F_n + F_0)\cdots)) \tag{3}$$

Where $H_{RBWA}^m(\cdot)$ represents the mth RBWA, $F_m$ represents the output features of the mth RBWA, and more information about RBWA will be introduced in 3.5.

Finally, $I_{SR}$ is generated by a convolution layer and bicubic interpolation algorithm:

$$I_{SR} = Conv_{3\times3}(F_m) + H_{up}(I_{LR}) \tag{4}$$

Here, $H_{up}(\cdot)$ represents the bicubic interpolation up-sampling operation.

**3.2 Space-Channel Adaptive Coordination Block(SCACB)**

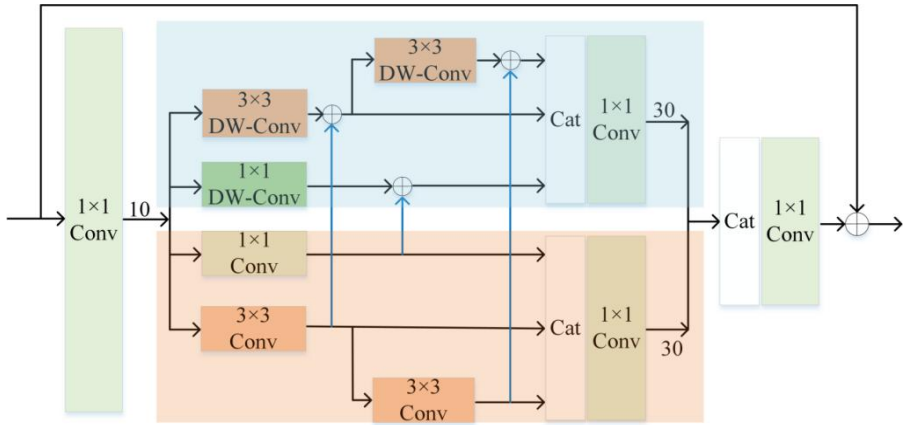

Fig. 2 Internal structure diagram of SCACB (3×3 Conv means ordinary convolution with 3×3 kernels, 3×3 DW-Conv means depthwise convolution with 3×3 kernels, and Cat means stitching operation by channel dimension)

SCACB mainly consists of ordinary convolution and depthwise convolution (DW-Con), which achieves adaptive adjustment of features in channels and space at the same time. First, the number of channels of features is compressed from 60 to 10 using 1×1 convolution layer to reduce the redundant information in it and realize the refinement of features. Then, the features are transported to the spatial, channel adaptive part, respectively (as shown in the light blue and light red parts in Figure 2). In the channel adaptation part, the network obtains information under three levels of perceptual fields, 1×1, 3×3, and 5×5, through three parallel branches. The network aggregates multiple channel information with the help of splicing and convolution layers to achieve adaptive adjustment on the channels. The spatial adaptive part is similar in structure to the channel adaptive part, but differs in that it uses depthwise convolution to obtain spatial information at different levels and realize the spatial adaptive adjustment of the network. More importantly, in the channel and spatial adaptive parts, we realize the information exchange in the corresponding stages by a simple jump connection (as shown by the blue line in Fig. 2), so that the two can work in a coordinated manner and let the network select the information of interest in space and channel autonomously, thus enhancing the representational ability of the network.

**3.3 Double Refinement Attention Communication Block (DRACB)**

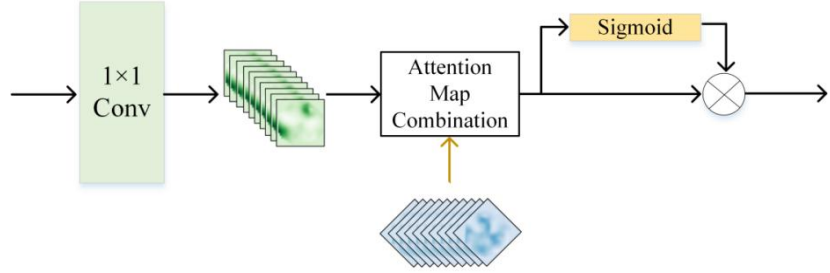

Fig. 3 Internal structure diagram of DRACB

DRACB achieves double refinement of features, and its internal structure is shown in Fig. 3. The 1 × 1 convolution layer has the dual functions of feature distillation and establishing pixel-level attention relationship. While compressing the input features, combined with the Sigmoid function, the pixel-level feature attention is also established, and the double refinement of the features is realized. It should be particularly noted that we introduce the attention maps generated by the previous DRACB (as shown in the blue part in Figure 3) to achieve the continuous flow of attention information in the attention module through a simple linear combination operation, maintaining the continuous guidance of the network to the information of interest. In this way, the previous attention map has a certain guiding effect on the current attention module.

The information exchange between attention modules is based on the communication between attention maps. Specifically, we use a linear combination to combine the attention maps in two attention modules, and then generate the attention masks by means of the Sigmoid function:

$$F^{'} = W * F_{l-1} + F_{l} \tag{5}$$

$$F = F^{'} * Sigmoid(F^{'}) \tag{6}$$

Where $F_{l-1}$ and $F_{l}$ denote the previous and current DRACB generated attention maps,

respectively, $F'$ denotes the fused attention maps, $W$ denotes the information exchange weight, which is set to 0.2 in this paper, and $F$ denotes the final generated features.

**3.4 Multilevel Refinement Enhancement block(MREB)**

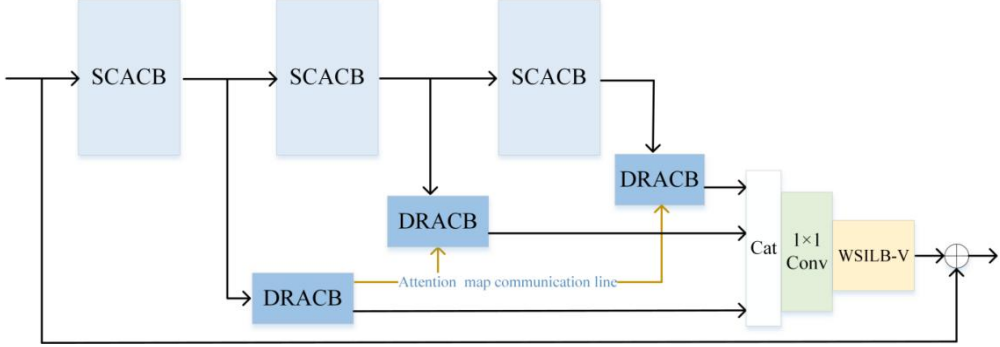

Fig. 4 Internal structure diagram of MREB (SCACB means Space-Channel Adaptive Coordination Block, DRACB means Double Refinement Attention Communication Block)

The internal structure of MREB is shown in figure 4, which is mainly composed of SCACB, DRACB, etc. The distillation mechanism of IMDN[23] and RFDN[24] can well retain the important features of each stage, so we also use this distillation mechanism, but it is different from them:

1. We use SCACB instead of SRB in RFDN. Compared with SRB with a single receptive field, SCACB can obtain spatial information and channel information under multiple receptive fields. Most importantly, SCACB has the adaptability of both space and channel, which allows the network to choose the information of interest at the channel and spatial level.

2. We use DRACB instead of 1 × 1 convolution layer in RFDN. The 1 × 1 convolution layer in RFDN only has the function of distillation and can not continuously and effectively guide the network to pay attention to the information of interest. The most special thing is that DRACB not only has the function of feature distillation, but also gives the network the ability to focus on sensitive pixels, realizing the double refinement of key features.

3. What is most unusual is that we have realized the information exchange of attention maps between adjacent DRACB. The attention modules in the previous advanced algorithms play a separate role, and there is a lack of effective information communication between the adjacent attention modules, which leads to the discontinuous flow of attention information within the module. In contrast, we realize the continuous flow of attention information between DRACB, realize the continuous guidance to the network, and let the network learn more powerful feature representation.

MREB processes features through multiple SCACB to obtain rich variety of hierarchical features at different stages. Then, we use DRACB to double refine the features of each stage, while with the help of the communication between attention maps, we can better guide the network to pay attention to key information. In addition, we use splicing and convolution layers to aggregate the important features at each stage. The WSILB-V used here is a variation of the WSILB(Weight-Sharing Information Lossless Attention Block) proposed in the paper [26], with the change that we use two 1×1 convolution layers instead of one 1×1 convolution layer to achieve feature compression and recovery, making it more lightweight.

### 3.5 Reconstruction Block With Attention(RBWA)

RBWA achieves feature size expansion while guiding the network to focus on the high-frequency information in it, and its internal structure is shown in Figure 1. First, the up-sampling operation of features is implemented using bilinear interpolation algorithm (BIA), and the inter-dependencies between features are strengthened by 3×3 convolution layers. Then, with the power of WSILB-V, the network is strengthened to focus on the high-frequency information in the features. Finally, the number of channels of features is compressed by convolution layers to reduce the parameter burden of the network.

## 4 Experiment

### 4.1 Datasets And Evaluation Metrics

We use the DIV2K [38] datasets containing a large number of high-quality images as our training set, and the low-resolution images needed for training are obtained by bilateral down-sampling. Testing uses Set5[39], Set14[40], B100[41], Urban100[42] benchmark test sets. We evaluate the reconstruction performance of the algorithm using SSIM and PSNR only on the Y channel in YCbCr space.

### 4.2 Experiment Setup Details

We use horizontal and vertical flip and other data enhancement methods to deal with the training set to enhance the robustness of the network. The randomly cropped image patch of size 192 × 192 from the training set is used as the input of the network, and the batch is set to 16. We use the widely used $L_1$ function as the loss function of the network with an initial learning rate of 0.0005, and the learning rate decays to half of the original rate after every 600 epochs, for a total of 3000 epochs of training. We use 6 MREB as the backbone network for feature processing. The activation function uses GELU and uses Adam to update the network gradient. The algorithm in this paper is based on Pytorch framework, and the experimental hardware platform is NVIDIA Tesla V100-PCIE-16GB.

### 4.3 Ablation study

We carried out experiments to verify the effectiveness of the important components of the network, and all the experimental results were obtained by training 900epochs under the same parameter setting.

Table 1 Impact of different number of MREBs on network performance

| Amount | 3 | 4 | 5 | 6 | 7 | 8 |
|---|---|---|---|---|---|---|
| PSNR(Set5) | 32.06 | 32.12 | 32.16 | 32.24 | 32.20 | 32.21 |
| SSIM(Set5) | 0.8920 | 0.8930 | 0.8932 | 0.8943 | 0.8942 | 0.8944 |
| Parameter | 196K | 230K | 264K | 298K | 332K | 366K |

**Ablation study of the number of MREB:** We studied the effect of different numbers of MREB on the network performance, and the experimental results are shown in Table 1. It is clear from the table that: the performance of the network increases with the increase of the number of MREB, but there is a small decrease in the performance of the network when the number of MREB reaches 7 or 8, in addition, there is a certain increase in the number of parameters. In summary, the optimal setting for the number of MREB is 6, which achieves optimal PSNR and sub-optimal SSIM results on Set5, and moderate number of parameters, in short, it achieves a good balance between the performance of the network and the number of parameters.

| Information Communication | 0 | 0.1 | 0.2 | 0.3 | 0.4 | 0.5 |
|---|---|---|---|---|---|---|
| PSNR(Set5) | 32.20 | 32.21 | 32.24 | 32.20 | 32.18 | 32.23 |
| SSIM(Set5) | 0.8941 | 0.8935 | 0.8943 | 0.8934 | 0.8938 | 0.8944 |
| Information Communication | 0.6 | 0.7 | 0.8 | 0.9 | 1.0 | |
| PSNR(Set5) | 32.21 | 32.14 | 32.19 | 32.20 | 32.18 | |
| SSIM(Set5) | 0.8940 | 0.8931 | 0.8940 | 0.8940 | 0.8937 | |

Table 2 Effect of the amount of information exchange between attention maps on network performance (0 in the table means no information exchange between attention maps, 0.1 means 0.1 information exchange between attention maps.)

**Effectiveness of communication between attention modules:** As shown in Table 2, we have conducted an experimental study on the amount of intentional communication between attention modules.

When the information communication is 0, that is, the attention module works alone, it achieves a good effect that the PSNR is 32.20dB and the SSIM is 0.8941. However, when the information is fully communicated, the amount of information exchange is 1, its performance decreases compared with the former to a certain extent. When the amount of information exchange is between 0 and 1, both PSNR and SSIM show some volatility. It can be seen that when the information communication is 0.2, the PSNR is the highest, reaching 32.24dB and SSIM is the second best. It can be seen that certain information communication between attention modules helps to improve the network performance. In other words, the former attention module provides guidance for the latter attention module through the communication between attention maps, which makes it easier for the network to learn more discriminative feature representation.

Table 3 Experimental studies on the effectiveness of SCACB(Only Spatial Adaptation (OSA) indicates that only the spatial adaptation part of SCACB is available. Only Channel Adaptation (OCA) means that there is only a channel adaptation part in SCACB. Also, for fairness, both OSA and OCA keep the same number of channels as SCACB. Space and Channel with No Communication (SCNC) means that the information exchange line between space and channel is removed from SCACB (as shown by the blue line in Figure 2).)

| Model | OSA | OCA | SCNC | SCACB(Our) |
|---|---|---|---|---|
| PSNR(Set5) | 32.15 | 32.22 | 32.19 | 32.24 |
| SSIM(Set5) | 0.8929 | 0.8948 | 0.8938 | 0.8943 |
| Parameters | 245K | 375K | 298K | 298K |

**Ablation study of SCACB:** SCACB achieves adaptive adjustment of features at both spatial and channel levels. To verify its effectiveness, we conducted a relevant experimental study, and the experimental results are shown in Table 3.

Compared with SCACB, OSA shows different degrees of decrease in both PSNR and SSIM, which indicates that the module only has spatial adaption does not allow the network to learn the information of interest well. For OCA, although its SSIM is higher than that of SCACB, its PSNR is lower than that of SCACB. More importantly, the number of parameters of OCA reaches 375 K, which is much higher than that of SCACB at 298 K. It can be seen that both single spatial and channel adaptive modules are difficult to achieve satisfactory results. When the module has both spatial and channel adaptive parts and both work independently, **i.e.** SCNC, both PSNR and SSIM of SCNC are in between the values of OSA and OCA. In addition, the PSNR and SSIM of SCNC are much lower than those of SCACB, which naturally leads to the conclusion that it is difficult to maximize the potential of the network with a single spatial and channel adaptive parts. In addition, when the network has both spatial and channel adaptation, it is important to exchange information with each other, and its can make the network work more efficiently and coordinately to further improve the reconstruction performance.

Table 4 Experimental study on the effectiveness of DRACB(We use schemes such as Distillation block, which is only composed of a 1 × 1 convolution layer, and information exchange between adjacent distillation blocks through jump connection on the basis of distillation blocks(Distillation block + jump connection) , instead of DRACB, respectively.)

| Model | PSNR(Set5) | SSIM(Set5) | Parameters |
|---|---|---|---|
| Distillation block | 32.16 | 0.8936 | 298K |
| Distillation block + Sigmoid | 32.20 | 0.8941 | 298K |
| Distillation block + jump connection | 32.15 | 0.8932 | 298K |
| Distillation block + jump connection + Sigmoid(DRACB) | 32.24 | 0.8943 | 298K |

**Ablation study of DRACB:** DRACB implements a double refinement of features and can guide the network to obtain the information of interest. To demonstrate its effectiveness, we conducted a relevant ablation study on it, the experimental results are shown in Table 4.

For the distillation block with only one refinement effect, the PSNR reached 32.16 dB and the SSIM reached only 0.8936, which is far inferior to that obtained by DRACB, while the number of parameters remained the same. Even if Sigmoid is added to the distillation block and attention mechanism is introduced, the results are hardly comparable to DRACB. In addition, DRACB achieves significant performance improvement without introducing additional number of parameters. The reason for the success of DRACB may be attributed not only to its ability to have feature distillation, but also to its ability to guide the network and focus on more critical information in the refined features. More importantly, it establishes a bridge for communication of attention map between attention modules and maintains a continuous flow of attention information, thus enabling a continuous guidance of the network.

### 4.4. Comparison with state-of-the-arts

To demonstrate that MREN has advanced reconstruction performance, we compare the results of MREN with some other advanced lightweight algorithms under ×2, ×3, and ×4 reconstruction tasks, as shown in Table 4, which include [SRCNN][4], [VDSR][5], [DRCN][11], [MemNet][6], [CARN][14], [IDN][22], [IMDN][23], [DNCL][1], [FilterNeL][2], [CFSRCNN][9], WMRN[19], MADNet[20], LMAN[21], [LAPAR-B][28], [HPUN-S][29], [ESRGCNN][3], [EFDN][25] etc.

Table 5 comparison of objective indicators with other advanced reconstruction algorithms

| Scale | Method | Year | parameters | Set5 | | Set14 | | B100 | | Urban100 | |
|---|---|---|---|---|---|---|---|---|---|---|---|
| | | | | PSNR | SSIM | PSNR | SSIM | PSNR | SSIM | PSNR | SSIM |
| ×2 | SRCNN | 2016 | 57K | 36.66 | 0.9542 | 32.42 | 0.9063 | 31.36 | 0.8879 | 29.50 | 0.8946 |
| | VDSR | 2016 | 665K | 37.53 | 0.9587 | 33.03 | 0.9124 | 31.90 | 0.8960 | 30.76 | 0.9140 |
| | DRCN | 2016 | 1774K | 37.63 | 0.9588 | 33.04 | 0.9118 | 31.85 | 0.8942 | 30.75 | 0.9133 |
| | MemNet | 2017 | 677K | 37.78 | 0.9597 | 33.28 | 0.9142 | 32.08 | 0.8978 | 31.31 | 0.9195 |
| | CARN | 2018 | 1592K | 37.76 | 0.9590 | 33.52 | 0.9166 | 32.09 | 0.8978 | 31.92 | 0.9256 |
| | IDN | 2018 | 579K | 37.83 | 0.9600 | 33.30 | 0.9148 | 32.08 | 0.8985 | 31.27 | 0.9196 |
| | IMDN | 2019 | 694K | 38.00 | 0.9605 | 33.63 | 0.9177 | 32.19 | 0.8996 | 32.17 | 0.9283 |
| | DNCL | 2019 | 694K | 37.65 | 0.9599 | 33.18 | 0.9141 | 31.97 | 0.8971 | 30.89 | 0.9158 |
| | FilterNeL | 2020 | 1374K | 37.86 | **0.9610** | 33.34 | 0.9150 | 32.09 | 0.8990 | 31.24 | 0.9200 |
| | CFSRCNN | 2020 | 1200K | 37.79 | 0.9591 | 33.51 | 0.9165 | 32.11 | 0.8988 | 32.07 | 0.9273 |
| | LAPAR-B | 2020 | 250K | 37.87 | 0.9600 | 33.39 | 0.9162 | 32.10 | 0.8987 | 31.62 | 0.9235 |
| | LMAN-s | 2021 | 525K | 37.94 | 0.9603 | 33.49 | 0.9167 | 32.08 | 0.8984 | 31.85 | 0.9251 |
| | WMRN | 2021 | 452K | 37.83 | 0.9599 | 33.41 | 0.9162 | 32.08 | 0.8984 | 31.68 | 0.9241 |
| | MADNet-$L_F$ | 2021 | 878K | 37.85 | 0.9600 | 33.39 | 0.9161 | 32.05 | 0.8981 | 31.59 | 0.9234 |
| | HPUN-S | 2022 | 226K | 37.93 | 0.9604 | 33.46 | 0.9170 | 32.13 | 0.8990 | 31.70 | 0.9245 |
| | ESRGCNN | 2022 | 1238K | 37.79 | 0.9589 | 33.48 | 0.9166 | 32.08 | 0.8978 | 32.02 | 0.9222 |
| | MREN | | 274K | 37.98 | 0.9597 | **33.67** | **0.9182** | 32.15 | 0.8986 | 32.10 | 0.9267 |
| | MREN-L | | 531K | **38.04** | 0.9599 | 33.65 | 0.9178 | **32.22** | **0.8997** | **32.35** | **0.9292** |
| ×3 | SRCNN | 2016 | 57K | 32.75 | 0.9090 | 29.28 | 0.8209 | 28.41 | 0.7863 | 26.24 | 0.7989 |
| | VDSR | 2016 | 665K | 33.66 | 0.9213 | 29.77 | 0.8314 | 28.82 | 0.7976 | 27.14 | 0.8279 |
| | DRCN | 2016 | 1774K | 33.82 | 0.9226 | 29.76 | 0.8311 | 28.80 | 0.7963 | 27.15 | 0.8276 |
| | MemNet | 2017 | 677K | 34.09 | 0.9248 | 30.00 | 0.8350 | 28.96 | 0.8001 | 27.56 | 0.8376 |
| | CARN | 2018 | 1592K | 34.29 | 0.9255 | 30.29 | 0.8407 | 29.06 | 0.8034 | 28.06 | 0.8493 |
| | IDN | 2018 | 588K | 34.11 | 0.9253 | 29.99 | 0.8354 | 28.95 | 0.8013 | 27.42 | 0.8359 |
| | IMDN | 2019 | 703K | 34.36 | 0.9270 | 30.32 | 0.8417 | 29.09 | 0.8046 | 28.17 | **0.8519** |
| | DNCL | 2019 | 694K | 33.95 | 0.9232 | 29.93 | 0.8340 | 28.91 | 0.7995 | 27.27 | 0.8326 |
| | FilterNeL | 2020 | 1374K | 34.08 | 0.9250 | 30.03 | 0.8370 | 28.95 | 0.8030 | 27.55 | 0.8380 |
| | CFSRCNN | 2020 | -- | 34.24 | 0.9256 | 30.27 | 0.8410 | 29.03 | 0.8035 | 28.04 | 0.8496 |
| | LAPAR-B | 2020 | 276K | 34.20 | 0.9256 | 30.17 | 0.8387 | 29.03 | 0.8032 | 27.85 | 0.8459 |
| | LMAN-s | 2021 | 709K | 34.31 | 0.9265 | 30.24 | 0.8397 | 29.02 | 0.8030 | 28.02 | 0.8487 |
| | WMRN | 2021 | 556K | 34.11 | 0.9251 | 30.17 | 0.8390 | 28.98 | 0.8021 | 27.80 | 0.8448 |
| | MADNet-$L_F$ | 2021 | 930K | 34.14 | 0.9251 | 30.20 | 0.8395 | 28.98 | 0.8023 | 27.78 | 0.8439 |
| | HPUN-S | 2022 | 234K | 34.32 | 0.9262 | 30.25 | 0.8404 | 29.05 | 0.8036 | 27.80 | 0.8456 |
| | ESRGCNN | 2022 | -- | 34.24 | 0.9252 | 30.29 | 0.8413 | 29.05 | 0.8036 | 27.14 | 0.8512 |
| | MREN | | 298K | 34.42 | 0.9265 | 30.30 | 0.8403 | 29.07 | 0.8039 | 28.15 | 0.8507 |
| | MREN-L | | 579K | **34.42** | **0.9265** | **30.36** | **0.8416** | **29.11** | **0.8046** | **28.19** | 0.8506 |
| ×4 | SRCNN | 2016 | 57K | 30.48 | 0.8628 | 27.49 | 0.7503 | 26.90 | 0.7101 | 24.52 | 0.7221 |
| | VDSR | 2016 | 665K | 31.35 | 0.8838 | 28.01 | 0.7674 | 27.29 | 0.7251 | 25.18 | 0.7524 |
| | DRCN | 2016 | 1774K | 31.53 | 0.8854 | 28.02 | 0.7670 | 27.23 | 0.7233 | 25.14 | 0.7510 |
| | MemNet | 2017 | 677K | 31.74 | 0.8893 | 28.26 | 0.7723 | 27.40 | 0.7281 | 25.50 | 0.7630 |
| | CARN | 2018 | 1592K | 32.13 | 0.8937 | 28.60 | 0.7806 | 27.58 | 0.7349 | 26.07 | 0.7837 |

| | | | | | | | | | | |
|---|---|---|---|---|---|---|---|---|---|---|
| | IDN | 2018 | 600K | 31.82 | 0.8903 | 28.25 | 0.7730 | 27.41 | 0.7297 | 25.41 | 0.7632 |
| | IMDN | 2019 | 715K | 32.21 | 0.8948 | 28.58 | 0.7811 | 27.56 | 0.7353 | 26.04 | 0.7838 |
| | DNCL | 2019 | 694K | 31.66 | 0.8871 | 28.23 | 0.7717 | 27.39 | 0.7282 | 25.36 | 0.7606 |
| | FilterNeL | 2020 | 1374K | 31.74 | 0.8900 | 28.27 | 0.7730 | 27.39 | 0.7290 | 25.53 | 0.7680 |
| | CFSRCNN | 2020 | -- | 32.06 | 0.8920 | 28.57 | 0.7800 | 27.53 | 0.7333 | 26.03 | 0.7824 |
| | LAPAR-B | 2020 | 313K | 31.94 | 0.8917 | 28.46 | 0.7784 | 27.52 | 0.7335 | 25.85 | 0.7772 |
| | LMAN-s | 2021 | 672K | 32.12 | 0.8939 | 28.53 | 0.7798 | 27.51 | 0.7340 | 25.96 | 0.7813 |
| | WMRN | 2021 | 536K | 32.00 | 0.8925 | 28.47 | 0.7786 | 27.49 | 0.7328 | 25.89 | 0.7789 |
| | MADNet-L$_F$ | 2021 | 1002K | 32.01 | 0.8925 | 28.45 | 0.7781 | 27.47 | 0.7327 | 25.77 | 0.7751 |
| | EFDN | 2022 | 276K | 32.08 | 0.8931 | 28.58 | 0.7809 | 27.56 | 0.7354 | 26.00 | 0.7815 |
| | HPUN-S | 2022 | 246K | 32.00 | 0.8921 | 28.50 | 0.7792 | 27.51 | 0.7339 | 25.79 | 0.7769 |
| | ESRGCNN | 2022 | -- | 32.02 | 0.8920 | 28.57 | 0.7801 | 27.57 | 0.7348 | 26.10 | 0.7850 |
| | MREN | | 298K | 32.24 | 0.8947 | 28.59 | 0.7808 | 27.57 | 0.7348 | 26.10 | 0.7846 |
| | MREN-L | | 579K | **32.33** | **0.8957** | **28.62** | **0.7815** | **27.59** | **0.7364** | **26.10** | **0.7857** |

**Quantitative comparison:** On several benchmark test sets, MREN-L achieved good reconstruction performance in almost all reconstruction tasks with 531 K parameters. The reconstruction effect of MREN-L is more excellent than that of IMDN, and the number of parameters is 136K less than that of IMDN, which may be partly due to the fact that MREN-L can not only achieve adaptive adjustment in space and channels, but also coordinate with each other through the exchange of information, making it easier for the network to learn interesting information.

In particular, MREN with network model parameters of only 298K has achieved better reconstruction results than most advanced algorithms in several reconstruction tasks. For the ×4 reconstruction task, MREN achieves a much better PSNR than HPUN-S with the same level of parameter count, and its PSNR on the Set5 test set is 0.33 dB higher than that of HPUN-S. For the Urban100 test set, which contains a large number of complex texture images, MREN outperforms HPUN-S by 0.0077 on SSIM. This allows MREN with only 298K parameters to be used on small removable devices with limited storage, and to achieve good reconstruction performance.

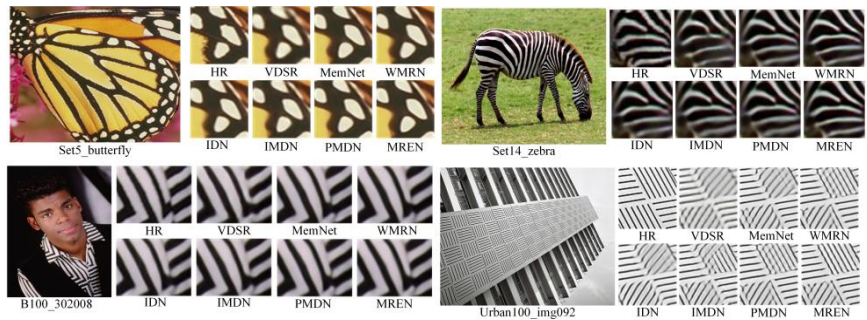

Figure 5 Visual comparison of ×4 reconstruction results on the test set

**Qualitative comparison：** In addition to performing objective metric comparisons, we also performed subjective visual comparisons, as shown in Figure 5. Compared with other advanced algorithms, our proposed MREN can reconstruct clearer and relatively richer texture results. For Urban100_img092, VDSR, MemNet, and IDN appear blurred to different degrees and have severe line distortion. Compared to IMDN, WMRN, MREN produces sharper results and there is less structural distortion.

Table 6 comparison of the complexity of different reconstructed networks under × 2 reconstruction tasks

| Method | Parameters | Flops | Set14 | | Urban100 | |
|---|---|---|---|---|---|---|
| IDN | 579K | 124.6G | 33.30 | 0.9148 | 31.27 | 0.9196 |
| IMDN | 694K | 158.8G | 33.63 | 0.9177 | 32.17 | 0.9283 |
| LAPAR-B | 250K | 85G | 33.39 | 0.9162 | 31.62 | 0.9235 |
| WMRN | 452K | 103G | 33.41 | 0.9162 | 31.68 | 0.9241 |
| MADNet-$L_F$ | 878K | 187.1G | 33.39 | 0.9161 | 31.59 | 0.9234 |
| HPUN-S | **226K** | 45.9G | 33.46 | 0.9170 | 31.70 | 0.9245 |
| MREN | 274K | **23.8G** | **33.67** | **0.9182** | 32.10 | 0.9267 |
| MREN-L | 531K | 46.3G | 33.65 | 0.9178 | **32.35** | **0.9292** |

**Computational complexity**：In addition, to evaluate the reconstruction performance of the model more comprehensively, we compared the number of parameters, computational complexity, PSNR and SSIM of the network, as shown in Table 6. It can be clearly seen from the table that MREN and MREN-L take up very little storage and computational resources while achieving relatively good objective metrics. In detail, MREN obtains better reconstruction performance with roughly the same number of parameters as HPUN-S. It is worth noting that its Flops is only 23.8G, which is about 52% of HPUN-S and 15% of IMDN. In addition, although the number of parameters of MREN-L reaches 531K, it is 347K less than MADNet-$L_F$ and 163K less than IMDN. At the same time, its Flops is much lower than MADNet-$L_F$ and IMDN. In this case, MREN-L achieves better reconstruction performance.

**5.Conclusion**

In this paper, we propose a lightweight spatial-channel adaptive coordination of multilevel refinement enhancement networks for image reconstruction. We construct a space-channel adaptive coordination block, which changes the previous situation that all features in the network operate independently on space and channel, and achieves interactive learning of the network at the space and channel levels. In addition, we also design a double refinement attention communication block, which achieves double refinement of features, and at the same time, achieves the communication of attention information between adjacent attentions by a simple linear combination operation, and maintains the continuous transmission of attention information, so as to achieve better continuous guidance of the network attention to the information of interest. Numerous experiments show that MREN achieves excellent results in terms of reconstruction results, number of network parameters, and computational complexity. In the future, we will use signal processing techniques and deep learning theory to design a more lightweight and low-complexity reconstruction network.

**References**

[1] Xie C, Zeng W, Lu X. Fast single-image super-resolution via deep network with component learning[J]. IEEE Transactions on Circuits and Systems for Video Technology, 2019, 29(12): 3473-3486.

[2] Li F, Bai H, Zhao Y. FilterNet: adaptive information filtering network for accurate and fast image super-resolution[J]. IEEE Transactions on Circuits and Systems for Video Technology, 2020, 30(6): 1511-1523.

[3] Tian C, Yuan Y, Zhang S, et al. Image Super-resolution with An Enhanced Group Convolutional Neural Network[J]. arXiv preprint arXiv:2205.14548, 2022.

[4] Dong C, Loy C C, He K, et al. Image Super-Resolution Using Deep Convolutional Networks[J]. IEEE Transactions on Pattern Analysis & Machine Intelligence, 2016, 38(02): 295-307.

[5] Kim J, Lee J K, Lee K M. Accurate image super-resolution using very deep convolutional networks[C]//Proceedings of the IEEE conference on computer vision and pattern recognition. 2016: 1646-1654.

[6] Tai Y, Yang J, Liu X, et al. Memnet: A persistent memory network for image restoration[C]//Proceedings of the IEEE international conference on computer vision. 2017: 4539-4547.

[7] Liu J, Zhang W, Tang Y, et al. Residual feature aggregation network for image super-resolution[C]//Proceedings of the IEEE/CVF conference on computer vision and pattern recognition. 2020: 2359-2368.

[8] Lim B, Son S, Kim H, et al. Enhanced deep residual networks for single image super-resolution[C]//Proceedings of the IEEE conference on computer vision and pattern recognition workshops. 2017: 136-144.

[9] Tian C, Xu Y, Zuo W, et al. Coarse-to-fine CNN for image super-resolution[J]. IEEE Transactions on Multimedia, 2020, 23: 1489-1502.

[10] Luo X, Xie Y, Zhang Y, et al. Latticenet: Towards lightweight image super-resolution with lattice block[C]//European Conference on Computer Vision. Springer, Cham, 2020: 272-289.

[11] Kim J, Lee J K, Lee K M. Deeply-recursive convolutional network for image super-resolution[C]//Proceedings of the IEEE conference on computer vision and pattern recognition. 2016: 1637-1645.

[12] Tai Y, Yang J, Liu X. Image super-resolution via deep recursive residual network[C]//Proceedings of the IEEE conference on computer vision and pattern recognition. 2017: 3147-3155.

[13] Wang L, Dong X, Wang Y, et al. Exploring sparsity in image super-resolution for efficient inference[C]//Proceedings of the IEEE/CVF conference on computer vision and pattern recognition. 2021: 4917-4926.

[14] Ahn N, Kang B, Sohn K A. Fast, accurate, and lightweight super-resolution with cascading residual network[C]//Proceedings of the European conference on computer vision (ECCV). 2018: 252-268.

[15] Lan R, Sun L, Liu Z, et al. MADNet: a fast and lightweight network for single-image super resolution[J]. IEEE transactions on cybernetics, 2020, 51(3): 1443-1453.

[16] He Z, Cao Y, Du L, et al. MRFN: Multi-receptive-field network for fast and accurate single

image super-resolution[J]. IEEE Transactions on Multimedia, 2020, 22(4): 1042-1054.
[17] Cai Y, Gao G, Jia Z, et al. Image Reconstruction of Multibranch Feature Multiplexing Fusion Network with Mixed Multilayer Attention[J]. Remote Sensing, 2022, 14(9): 2029.
[18] Liu Y, Jia Q, Fan X, et al. Cross-srn: Structure-preserving super-resolution network with cross convolution[J]. IEEE Transactions on Circuits and Systems for Video Technology, 2021.
[19] Sun L, Liu Z, Sun X, et al. Lightweight image super-resolution via weighted multi-scale residual network[J]. IEEE/CAA Journal of Automatica Sinica, 2021, 8(7): 1271-1280.
[20] Lan R, Sun L, Liu Z, et al. MADNet: a fast and lightweight network for single-image super resolution[J]. IEEE transactions on cybernetics, 2021, 51(3): 1443-1453.
[21] Wan J, Yin H, Liu Z, et al. Lightweight image super-resolution by multi-scale aggregation[J]. IEEE Transactions on Broadcasting, 2021, 67(2): 372-382.
[22] Hui Z, Wang X, Gao X. Fast and accurate single image super-resolution via information distillation network[C]//Proceedings of the IEEE conference on computer vision and pattern recognition. 2018: 723-731.
[23] Hui Z, Gao X, Yang Y, et al. Lightweight image super-resolution with information multi-distillation network[C]//Proceedings of the 27th acm international conference on multimedia. 2019: 2024-2032.
[24] Liu J, Tang J, Wu G. Residual feature distillation network for lightweight image super-resolution[C]//European Conference on Computer Vision. Springer, Cham, 2020: 41-55.
[25] Wang Y. Edge-enhanced Feature Distillation Network for Efficient Super-Resolution[C]//Proceedings of the IEEE/CVF Conference on Computer Vision and Pattern Recognition. 2022: 777-785.
[26] Cai Y, Gao G, Jia Z, et al. Image Reconstruction Based on Progressive Multistage Distillation Convolution Neural Network[J]. Computational Intelligence and Neuroscience, 2022, 2022.
[27] Gao G, Li W, Li J, et al. Feature Distillation Interaction Weighting Network for Lightweight Image Super-Resolution[J]. arXiv preprint arXiv:2112.08655, 2021.
[28] Li W, Zhou K, Qi L, et al. Lapar: Linearly-assembled pixel-adaptive regression network for single image super-resolution and beyond[J]. Advances in Neural Information Processing Systems, 2020, 33: 20343-20355.
[29] Sun B, Zhang Y, Jiang S, et al. Hybrid Pixel-Unshuffled Network for Lightweight Image Super-Resolution[J]. arXiv preprint arXiv:2203.08921, 2022.
[30] Tan C, Cheng S, Wang L. Efficient image super-resolution via self-calibrated feature fuse[J]. Sensors, 2022, 22(1): 329.
[31] Hu J, Shen L, Sun G. Squeeze-and-excitation networks[C]//Proceedings of the IEEE conference on computer vision and pattern recognition. 2018: 7132-7141.
[32] Hou Q, Zhou D, Feng J. Coordinate attention for efficient mobile network design[C]//Proceedings of the IEEE/CVF conference on computer vision and pattern recognition. 2021: 13713-13722.
[33] Niu B, Wen W, Ren W, et al. Single image super-resolution via a holistic attention network[C]//European conference on computer vision. Springer, Cham, 2020: 191-207.
[34] Wang Z, Gao G, Li J, et al. Lightweight image super-resolution with multi-scale feature interaction network[C]//2021 IEEE International Conference on Multimedia and Expo (ICME). IEEE, 2021: 1-6.

[35] Zhao H, Kong X, He J, et al. Efficient image super-resolution using pixel attention[C]//European Conference on Computer Vision. Springer, Cham, 2020: 56-72.
[36] Ledig C, Theis L, Huszár F, et al. Photo-realistic single image super-resolution using a generative adversarial network[C]//Proceedings of the IEEE conference on computer vision and pattern recognition. 2017: 4681-4690.
[37] Lim B, Son S, Kim H, et al. Enhanced deep residual networks for single image super-resolution[C]//Proceedings of the IEEE conference on computer vision and pattern recognition workshops. 2017: 136-144.
[38] Timofte R, Agustsson E, Van Gool L, et al. Ntire 2017 challenge on single image super-resolution: Methods and results[C]//Proceedings of the IEEE conference on computer vision and pattern recognition workshops. 2017: 114-125.
[39] Bevilacqua M, Roumy A, Guillemot C, et al. Low-complexity single-image super-resolution based on nonnegative neighbor embedding[J]. 2012.
[40] Zeyde R, Elad M, Protter M. On single image scale-up using sparse-representations[C]//International conference on curves and surfaces. Springer, Berlin, Heidelberg, 2010: 711-730.
[41] Martin D, Fowlkes C, Tal D, et al. A database of human segmented natural images and its application to evaluating segmentation algorithms and measuring ecological statistics[C]//Proceedings Eighth IEEE International Conference on Computer Vision. ICCV 2001. IEEE, 2001, 2: 416-423.
[42] Huang J B, Singh A, Ahuja N. Single image super-resolution from transformed self-exemplars[C]//Proceedings of the IEEE conference on computer vision and pattern recognition. 2015: 5197-5206.